\documentclass[a4paper,UKenglish,cleveref, autoref, thm-restate]{article}

\title{User Preferences and the Shortest Path} 

\author{Isabella Kreller and Bernd Ludwig\\University of Regensburg,\\Universitätsstraße 31\\D-93053 Regensburg, Germany\\({\tt bernd.ludwig@ur.de)}}

\usepackage[utf8]{inputenc}
\usepackage[ruled,vlined]{algorithm2e}
\usepackage{amsmath}
\usepackage{hyperref}
\usepackage{graphicx}
\usepackage{subcaption}
\DeclareMathOperator*{\argmax}{arg\,max}
\DeclareMathOperator*{\argmin}{arg\,min}
\begin{document}

\maketitle

\begin{abstract}
Indoor navigation systems leverage shortest path algorithms to calculate routes. In order to define the “shortest path”, a cost function has to be specified based on theories and heuristics in the application domain. For the domain of indoor routing, we survey theories and criteria identified in the literature as essential for human path planning. We drive quantitative definitions and integrate them into a cost function that weights each of the criteria separately. We then apply an exhaustive grid search to find weights that lead to an ideal cost function. “Ideal” here is defined as guiding the algorithm to plan routes that are most similar to those chosen by humans. To explore which criteria should be taken into account in an improved pathfinding algorithm, eleven different factors whose favorable impact on route selection has been established in past research were considered. Each factor was included separately in the Dijkstra algorithm and the similarity of thus calculated routes to the actual routes chosen by students at the University of Regensburg was determined. This allows for a quantitative assessment of the factors’ impact and further constitutes a way to directly compare them.  A reduction of the number of turns, streets, revolving doors, entryways, elevators as well as the combination of the aforementioned factors was found to have a positive effect and generate paths that were favored over the shortest path. Turns and the combination of criteria turned out to be most impactful.
\end{abstract}

\section{Introduction}

Finding a way through their immediate environment to reach a sought destination is a task highly relevant to humans since they see themselves confronted with it regularly. Due to the available options of traveling and visiting unfamiliar cities in foreign countries, the cultural differences as well as the possible language barrier may further complicate wayfinding tasks. Being able to facilitate situations like these, location-based navigation services, particularly those intended for car navigation, enjoy great popularity and have established themselves as indispensable supporting tools for navigation in the modern world, as indicated by the high usage of Google Maps \cite{haria2019working}.

Although still overshadowed by the ubiquity of outdoor navigation systems, indoor wayfinding is not anymore an obscure area of research and it might be that in this setting navigational support might be needed most.
Wayfinding, meaning the goal-oriented process of a person finding a predefined destination \cite{allen1999cognitive}, often involving navigation through large areas without initial perception of the target location \cite{kuipers1978modeling}, is deemed a cognitive challenging task requiring a variety of mental resources \cite{montello2013functions}.\\
\indent
In addition, indoor spaces exhibit structural attributes, such as the additional dimensionality inherent in multi-storied buildings \cite{holscher2006up} or the lack of reliable methods for localization \cite{sakpere2017state}, which give rise to further difficulties and require current systems to be adapted to this navigational context. \\
\indent
This likewise concerns the pathfinding algorithm: due to the current lack of geospatial research regarding this setting, employed algorithms have shown little to no deviation from optimizing only in regard to the route’s efficiency disregarding human preference. Previous studies, as discussed in subsections 2.2, 2.3, and 3.3, have determined a multitude of diverse criteria as influential to people’s path choices, yet quantification of their impact and applicability in indoor environments remain to be studied.

\section{Related Work}
\label{sec:typesetting-summary}

\subsection{Difficulties of Indoor Navigation}

The reasons for which indoor navigation systems lack the popularity and widespread use their outdoor counterparts are enjoying are twofold.\\ 
\indent 
Firstly, there exist additional difficulties regarding the implementation of indoor navigation systems. The additional third dimension, the height, inherent in multi-level buildings creates challenges regarding a suitable design for the visual projection onto the small two-dimensional screens of smartphones \cite{brunner2005active}. This is important for providing easily comprehensible and unambiguous instructions to the user to direct them to the correct floor. Hence, conventional visualizations and user interfaces of map-based applications are not predestined for wayfinding support inside buildings.\\ 
\indent 
Yet the most crucial obstacle to overcome is likely the absence of a reliable method of localization. Whereas the vast majority of outdoor navigation systems make use of Global Positioning System (GPS) signals \cite{sakpere2017state}, this does not constitute a viable option indoors. These signals manage to pass through the walls and ceilings of buildings only sporadically, rendering it an unsatisfactory solution for optimal navigational support \cite{sakpere2017state}. Although numerous alternatives, including WiFi, Bluetooth, and Infrared, have been proposed, these too entail drawbacks regarding performance, cost, or reliability. \cite{sakpere2017state}\\ 
\indent 
Furthermore, human orientation may also prove more challenging in this setting. The third dimension, i.e. multiple stories inside buildings, creates challenges not only for the system’s ability to display information but for human wayfinding as well. Staircases have been identified as the location where mistakes are being made most frequently during a wayfinding task \cite{holscher2006up}; other research supports the idea that people struggle with vertical, rather than horizontal, movement, which might be explained by the mistaken assumption that a building’s layout is identical across floors \cite{soeda1997wayfinding}. A restricted field of vision due to walls and corners makes the usage of landmarks as points of reference only possible if they are local or in direct line of view, whereas in outdoor spaces global landmarks, such as towers or the city skyline, can also support wayfinding. \cite{yang2011similarities}.

\subsection{Influential Factors}
There is reason to contest and doubt that humans' behavior is purely rational and can be explained with conscious and deliberate reasoning alone for we are susceptible to numerous biases in our ways of thinking and decision making \cite{kahneman2003maps}. 

The same holds for wayfinding behavior: while the most logical path to the desired destination would generally be the shortest or least time consuming one, this does not align with actual human conduct or preference \cite{armeni2013pedestrian, golledge1995path, armeni2013pedestrian, zhu2015people}.\\
\indent
Afyouni, Ray, and Claramunt \cite{afyouni2012spatial} suggest a classification of the context in which decision making takes place that can also be applied to wayfinding tasks. The user-centered context takes into account the navigator’s profile, such as mental capabilities of visualization \cite{sternberg1990metaphors}, or rotation of objects \cite{mayer1994visual}, their age \cite{taillade2016age} and gender \cite{lawton2001gender}, or cultural background \cite{hund2012impact}. The context of execution, on the other hand, describes qualities of the information systems which aid the user; the representation of maps and real-world references used by mobile applications \cite{puikkonen2009towards} has been found to impact navigational success, and so do conventional means of support, such as signs and paper maps to varying degrees. \\
\indent
What is most relevant for the adjustment of the pathfinding algorithm, however, can be described by the environmental context. \\
\indent
According to Lynch \cite{lynch1960image}, the components most crucial for the formation of mental maps are paths, edges, districts, and landmarks, which, belonging to the same mental models, exist in connection and mutual interference. This concept fits well into the research conducted, which suggests that humans mentally visualize their environment as being structurally equivalent to a network of streets and paths \cite{gillner1998navigation, kuipers1978modeling, meilinger2008working}.\\
\indent
In this sense, the definition of a path’s complexity contrived by Golledge \cite{golledge1995path} and since adopted into the work of many others \cite{duckham2003simplest, holscher2011would, richter2002you} is to consider the number of streets or turns connecting them. \\
\indent
Alternative approaches \cite{brown2013exploring, giannopoulos2014wayfinding, grum2005risk, richter2002you, vanclooster2013analyzing} of achieving simplicity take into account the number of decision points, i.e. intersections of multiple streets, or their branching factor.  It is defined by the number of streets meeting, thus considered a measure of difficulty or complexity of a decision point \cite{o1992effects} and increasing the possibility of losing one’s way \cite{allen1999cognitive}. \\
\indent
Rather than the number of branches, Dalton \cite{dalton2003secret} found a preference for choosing the maximum, rather than mean or minimum, angles of paths available at decision points, meaning that people were more likely to choose the way that was aligned most with from where they were coming, presumably as a complexity-minimizing strategy. Further observations support the idea of people’s predilection towards maintaining linearity in their routes \cite{conroy2001spatial, montello1991spatial, sadalla1989remembering, zacharias2001pedestrian}. Similarly, another study points towards a human preference of decreasing the angular distance to the destination (Turner, 2009) at decision points, favoring this over a reduction of the Manhattan distance. \\
\indent
Other than a preference for simple paths, the absence or presence of certain path entities could serve as a decisive factor. Landmarks, defined by their conspicuity and recognizability \cite{caduff2008assessment, lynch1960image} commonly function as points of reference and may help reduce cognitive load for orientation \cite{goodman2004using}. For staircases, both a preference \cite{kikiras2006ontology} and avoidance \cite{kikiras2006ontology, koide20053} have been identified.

\subsection{Alternatives to the Shortest Path}
Variants for the cost function in path planning have been proposed with different objectives in mind. Grum \cite{grum2005risk} describes a way of extending the Dijkstra algorithm so that it incorporates a weight for the risk of making a wrong decision at an intersection proportional to the distance of the mistakenly chosen path, assuming that the navigator notices their mistake at the next decision point and retraces their way. In this manner, not the shortest path, but the one with the least danger of getting lost is computed. Yet, the question remains open how the risk for wrong decisions may be estimated for arbitrary environments. \\
\indent
Duckham and Kulik \cite{duckham2003simplest} proposed a way of calculating the simplest path, using solely a measure for ease of navigation by weighting route instructions without taking distance into account at all. Instructions needed to navigate decision points were penalized according to adjusted slot values for turns and branches as proposed by Mark \cite{mark1986automated}, yielding the sought-after advantages. \\
\indent
Physically constrained people constitute a group of users which might benefit immensely from adjustments to conventional navigation systems. A proposed way of achieving this is by taking into account edge weights other than distances, thus considering specific user preferences such as avoiding stairs or minimizing turns which can be input directly by the user \cite{mantha2020evaluation}. \\
\indent
In contrast, introducing a measure of beauty as a way of suggesting short routes that are perceived to be emotionally pleasant was accomplished by crowdsourcing of people’s judgments of places along the dimensions beautiful, quiet, and happy, making use of pictures from Google Street View or taken by volunteers \cite{quercia2014shortest}. Such an aesthetic criterion has not been considered by the persons who provided us with their preferred routes, because they tried to reach their destination as quickly and easily as possible. However, \cite{doi:10.1080/13658816.2010.510799} report that minimizing the number of turns is a key criterion for route planning in outdoor environments. Such routes seems to be more easily perceivable and cognitively less demanding. 

In our study we evaluate, whether this result can be transferred to indoor environments and whether other criteria (e.g.\ level changes or use of elevators) that result from indoor architectural constraints also influence human decision making for route planning.

\section{Route Calculation}
To operationalize the above research question, we make use of the indoor navigation system URWalking \cite{muller2017path,epub43697} (\url{urwalking.ur.de}). It implements
{\sc Dijkstra}'s algorithm \cite{dijkstra1959note} to determine the shortest path in terms of the distance between starting point and destination in meters.

\subsection{User Preferred Routes}

Our empirical study is based on a data set of 221 routes, collected both during the summer (100 paths) and winter semester (121 paths) at the University of Regensburg, thus accommodating for differences in weather and temperature. All participants, 129 of which were male, stated to be highly familiar with the environment and ranged in age from 19 to 33 (mean age = 22.93, SD = 2.67). As means for data collection, participants were asked to walk and describe their daily route on the campus. The university campus spans over an expanse of 1 km² and includes 16 buildings with a sum of 86 individual floors and 5228 rooms, encompassing both paths indoors and outdoors \cite{muller2017path}.

The aim of the data collection was to find out whether modifications of URWalking's path planning algorithm could lead to routes users enjoy taking, so as to find a compromise between the theoretically minimal walking distance and the distance of preferred paths (i.e.\ can we approximately simulate human decision making using {\sc Dijkstra}'s efficient algorithm?). For the envisioned modifications, we identified criteria in research literature that are argued to impact human decision making during wayfinding tasks.

\subsection{Criteria}

The criteria that will be described in the following were chosen due to the extent of the research they are supported by and applied in, and also based on the feasibility of their incorporation into the {\sc Dijkstra} algorithm. The scope of literature on the measurement of path complexity suggests that there is no agreed-upon answer as to how to define it. Although a few studies have suggested more elaborate ways of ruling how complex a path is, a few elementary classifications resurface frequently across research papers. We identified 11 factors:

A straightforward way to determine complexity is by considering the \textbf{(1) number of turns} a route contains. Regarding the definition of what constitutes a turn in navigation the answers once again are ambiguous. According to the Cambridge Dictionary \cite{turn}, a turn can be considered "a change in the direction in which you are moving or facing". Although in some studies a sufficient change in directions achieved only when the angle between two path segments is smaller or equal to 90 degrees \cite{holscher2011would, vanclooster2019turn}, others interpret every branch at a decision point to count as a turn \cite{duckham2003simplest, klippel2005wayfinding}. 

A related factor influential to wayfinding is the \textbf{(2) number of streets}, meaning the path segments separated by decision points or turns of any kind \cite{holscher2011would}. Considering only turns up to a maximum angle of 90 degrees would not account for the number of streets whereas disregarding the angle entirely would run the risk of accounting for turns that might not be perceived as such due to their small deviation from a linear path. For these reasons, both of the definitions of what constitutes a turn were examined: weighting turns regardless of their angle will acknowledge the number of streets more completely, and following the definition in accordance with  \cite{holscher2011would} comprises a more traditional way of regarding turns and assessing their impact independently.

The \textbf{(3) number of decision points} themselves was considered for route calculation as well. Since the number of paths at a junction demanding a decision is two in addition to the path the navigator came from, a decision point in this study was defined as an intersection of three or more paths; their exact number was disregarded and every path located at a point which fits this definition was penalized in the same way. 

The \textbf{(4) branching factor}, meaning the number of intersecting paths, was regarded separately. To weigh the paths proportionally to their quantity, every path at a decision point received the additional weight times the number of intersecting paths. These weights were taken into account only at places that qualify as decision points corresponding to the aforementioned definition.

The findings on which paths at intersections constitute the most likely choice are ambiguous, which is why two divergent findings were considered. One is the preference for selecting the path which most closely aligns with the global direction of the destination \cite{turner2009role}, being equal to the avoidance of all paths other than the one with the \textbf{(5) minimum deviation angle}. 

Alternatively, a preference for \textbf{(6) linearity} \cite{conroy2001spatial} might influence the person navigating to continue their route along the most linear path, i.e. the path with the largest angle (between 0 and 180 degrees) in regard to the direction from which the navigator arrived. For this, all except the most linear branch at a decision point were penalized, allowing for the option of exempting two branches in case of identical largest angles. To keep with the definition of linearity as well as Dalton’s study \cite{dalton2003secret}, the condition was added that the largest angle had to span 150 degrees or more. If there was no such angle, e.g. in a T-crossing, no weighting was carried out.

Due to the contradictory nature of the literature regarding the preference for \textbf{(7) staircases} in routes, both a preference for and avoidance of stairs were considered. \textbf{(8) Elevators} constituted the only alternative for changing floors at the university; hence the avoidance of stairs, being equal to a preference for elevators,  was implemented by adding weights for every staircase and vice versa.

Another element to be taken into consideration for indoor environments exclusively is doors. The path network used for calculating routes includes different types of nodes that differentiate between \textbf{(9) doorways}, \textbf{(10) entrances} to closed premises such as offices or lecture halls, termed entryways, and \textbf{(11) revolving doors}. Since it has been found that doors take on the function of important landmarks in indoor spaces, and serve as transitions between separate spaces \cite{tian2013toward}, the effect of doors, with the above-mentioned differentiation, will be included in the analysis.

We modified the original URWalking path finding algorithm such as to separately take the 11 above factors into account and to identify  their optimal weights for calculating optimal routes. This constitutes a method of path calculation that, while still maintaining the objective of minimizing distance, can easily integrate different factors and assign them distinct levels of importance.

\begin{algorithm}
\SetAlgoLined
\KwResult{Shortest paths to all nodes starting from {\tt source}}
create vertex set Q\;
 \For{each vertex v in Graph}{%
dist[v] ← INFINITY\;                 
prev[v] ← UNDEFINED\;
 add v to Q\;
 dist[source] ← 0\;}

\While{Q is not empty}{%
u ← vertex in Q with min dist[u]\;
remove u from Q\;
\For{each neighbor v of u}{%
alt ← dist[u] + wlength(u, v)\;
\If{alt < dist[v]}{%
dist[v] ← alt\;
prev[v] ← u\;}}}
return dist[], prev[]\;
\caption{{\sc Dijkstra}'s algorithm (source: 
\href{https://en.wikipedia.org/wiki/Dijkstra's_algorithm}{Wikipedia})}
\label{code:da}
\end{algorithm}

The standard shortest path definition of $\mbox{wlength}(x,y)$ is the metric distance $\mbox{length}(x,y)$ between two nodes $x$ and $y$ connected by an edge. The operationalization of the identified factors is quite straightforward: we simply modified the cost function used by {\sc Dijkstra}'s algorithm (see Algorithm \ref{code:da}).  The key to taking the factors into account is a modification of the function $\mbox{length}(x,y)$ for the cost of edges in the graph. In the standard version, the function computes the distance between $x$ and $y$ in meters. In our modifications, we add artificial, heuristic costs to $\mbox{length}(x,y)$. We apply the idea that a factor can be accounted for by extending the distance between $x$ and $y$, i.e.\ by making artificially the edge $(x,y$) more expensive. For each factor, a particular criterion determines whether it actually applies for an edge currently considered by {\sc Dijkstra}'s algorithm:

\begin{enumerate}
    \item number of turns:
    
    $
    \mbox{wlength}(x,y,w)=\left\{
    \begin{array}{ll}
    \mbox{length}(x,y) + w & \mbox{angle}(x,y)\leq 90\\
     \mbox{length}(x,y) & \mbox{otherwise}\\
    \end{array}
    \right.
    $
    
\item number of streets:

    $
    \mbox{wlength}(x,y,w)=\left\{
    \begin{array}{ll}
    \mbox{length}(x,y) + w & \mbox{angle}(x,y)\leq 180\\
    &(\mbox{route does not continue straight on after $y$})\\
     \mbox{length}(x,y) & \mbox{otherwise}\\
    \end{array}
    \right.
    $
\item number of decision points:

    $
    \mbox{wlength}(x,y,w)=\left\{
    \begin{array}{ll}
    \mbox{length}(x,y) + w & \mbox{$x$ has at least three neighbours}\\
     \mbox{length}(x,y) & \mbox{otherwise}\\
    \end{array}
    \right.
    $
    
\item branching factor:

    $
    \mbox{wlength}(x,y,w)=\left\{
    \begin{array}{ll}
    \mbox{length}(x,y) + n\cdot weight & \mbox{$x$ has exactly $n$ neighbours and $n\geq 3$}\\
     \mbox{length}(x,y) & \mbox{otherwise}\\
    \end{array}
    \right.
    $
    
\item minimum deviation angle:

    $
    \mbox{wlength}(x,y,w)=\left\{
    \begin{array}{ll}
    \mbox{length}(x,y) + w & \mbox{$x$ has at least three neighbours}\\
    & \mbox{and $y\not=\argmin_z\mbox{angle}(x,z)$}\\
     \mbox{length}(x,y) & \mbox{otherwise}\\
    \end{array}
    \right.
    $
    
\item linearity:

    $
    \mbox{wlength}(x,y,w)=\left\{
    \begin{array}{ll}
    \mbox{length}(x,y) + w & \mbox{$x$ has at least three neighbours}\\
    & \mbox{and $y\not=\argmax_z\mbox{angle}(x,z)$}\\
    & \mbox{and $\max_z\mbox{angle}(x,z)\geq 150$}\\
     \mbox{length}(x,y) & \mbox{otherwise}\\
    \end{array}
    \right.
    $
\item staircases:

    $
    \mbox{wlength}(x,y,w)=\left\{
    \begin{array}{ll}
    \mbox{length}(x,y) + w & \mbox{$y$ is an edge of type {\tt staircase}}\\
     \mbox{length}(x,y) & \mbox{otherwise}\\
    \end{array}
    \right.
    $

\item elevators:

    $
    \mbox{wlength}(x,y,w)=\left\{
    \begin{array}{ll}
    \mbox{length}(x,y) + w & \mbox{$y$ is an edge of type {\tt elevator}}\\
     \mbox{length}(x,y) & \mbox{otherwise}\\
    \end{array}
    \right.
    $
    
\item doorways:

    $
    \mbox{wlength}(x,y,w)=\left\{
    \begin{array}{ll}
    \mbox{length}(x,y) + w & \mbox{$y$ is of type {\tt doorway}}\\
     \mbox{length}(x,y) & \mbox{otherwise}\\
    \end{array}
    \right.
    $
    
\item entrances:

    $
    \mbox{wlength}(x,y,w)=\left\{
    \begin{array}{ll}
    \mbox{length}(x,y) + w & \mbox{$y$ is of type {\tt entrance}}\\
     \mbox{length}(x,y) & \mbox{otherwise}\\
    \end{array}
    \right.
    $
    
\item revolving doors:

    $
    \mbox{wlength}(x,y,w)=\left\{
    \begin{array}{ll}
    \mbox{length}(x,y) + w & \mbox{$y$ is of type {\tt revolving-door}}\\
     \mbox{length}(x,y) & \mbox{otherwise}\\
    \end{array}
    \right.
    $
\end{enumerate}

In these definition $w$ is a constant whose value determines by how many meters the length of an edge between $x$ and $y$ is extended. Consequentially, the shortest path gets "longer" the more instances of the respective considered factor it includes.

With this model, our main task is now to determine values of $w$ for each factor that maximizes the similarity between routes calculated by URWalking and those chosen by the participants of our study.

\subsection{Systematic Search for Optimal Weights}

Our intuition is that a meaningful assessment of \textit{similarity between two routes} $r_1$ and $r_2$ reflects the length of shared path segments, i.e.\ edges in $r_1$ and $r_2$:
$$
\mbox{similarity}(r_1,r_2)=\frac{\displaystyle \sum_{e\in r_1\wedge e\in r_2}\mbox{length}(e)}{\min(\mbox{length}(r_1),\mbox{length}(r_2))}
$$
with $\mbox{length}(r)=\sum_{e\in r}\mbox{length}(e)$ for any route $r$ and $\mbox{length}(e)=\mbox{length}(x,y)$ with $x$ being the source node and $y$ the sink node of any edge $e$.

In other words, the calculated Similarity Score $\mbox{similarity}(r_1,r_2)$ was defined as the sum of the distances belonging to shared edges divided by the length of the shorter path (analogous to \cite{jan2000using}). Following Delling et al.’s \cite{delling2015navigation} approach, the mean of the similarity scores across all path comparisons was then considered for assessment of the algorithm, i.e. each factor’s individual impact on wayfinding.

To identify an optimal weight $w$ for each factor, we performed a systematic grid search for this one-dimensional parameter. Overall, we performed 11 independent searches considering a different factor in each search.

In such a search, we first set $w=0$ to identify the shortest path in terms of the metric distance between a fixed start and destination. Then we iterated over $w$ and incremented it by 1 in each iteration. As the cost function $\mbox{wlength}(x,y,w)$ computes costs in meters, incrementing $w$ by $1$ adds one meter to the physical distance between $x$ and $y$ allowing a detour of $1$ meter for a path still to be identified as the shortest path, this time however taking one of the identified factors into account. After having calculated this shortest path for a fixed $w$ we computed the similarity score to preferred routes in our data set with identical start and destination resulting in a list of similarity scores for which we could easily calculate a mean similarity. We repeated this process for all pairs $(\mbox{start},\mbox{destination})$ in our data set and calculated a mean similarity score for the fixed $w$.

\begin{figure}[tb]
\centering
    \begin{subfigure}{.5\textwidth}
        \includegraphics[width=0.95\linewidth]{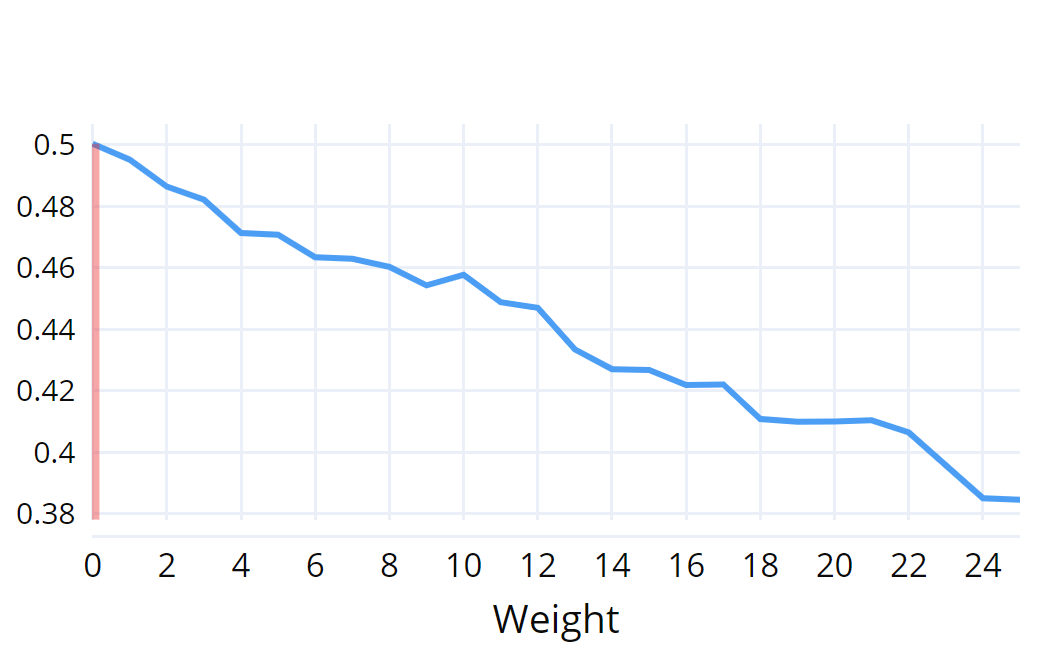}
        \caption{Similarity Scores of Angles}
        \label{fig:sub1}
    \end{subfigure}%
    \begin{subfigure}{.5\textwidth}
        \includegraphics[width=0.95\linewidth]{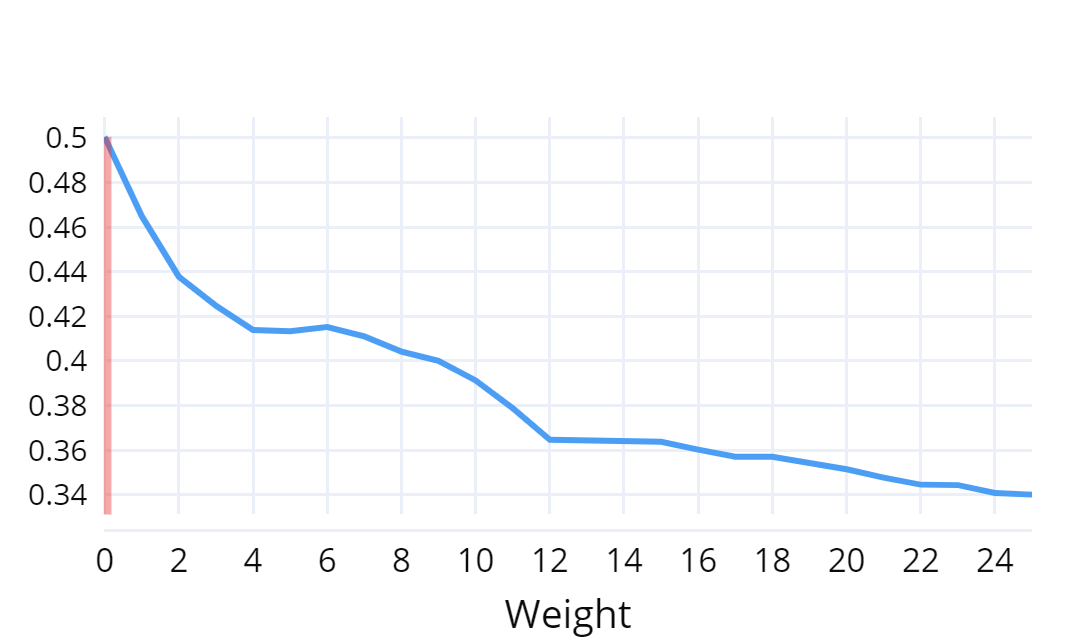}
        \caption{Similarity Scores of Branching Factor}
        \label{fig:sub2}
    \end{subfigure}
    \begin{subfigure}{.5\textwidth}
        \includegraphics[width=0.95\linewidth]{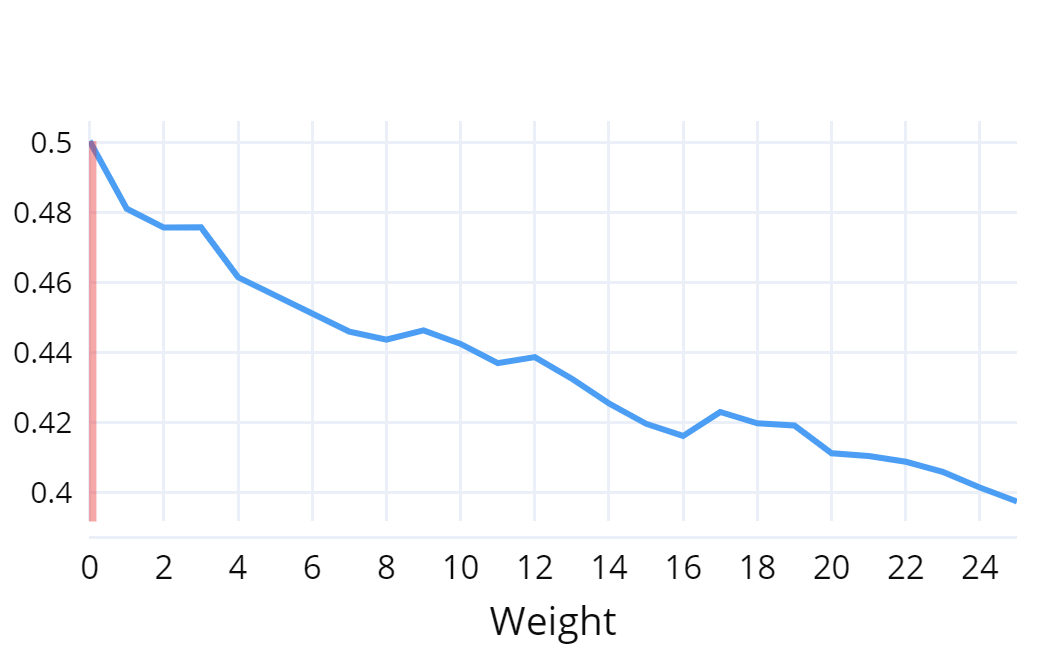}
        \caption{Similarity Scores of Decision Points}
        \label{fig:sub3}
    \end{subfigure}%
    \begin{subfigure}{.5\textwidth}
        \includegraphics[width=0.95\linewidth]{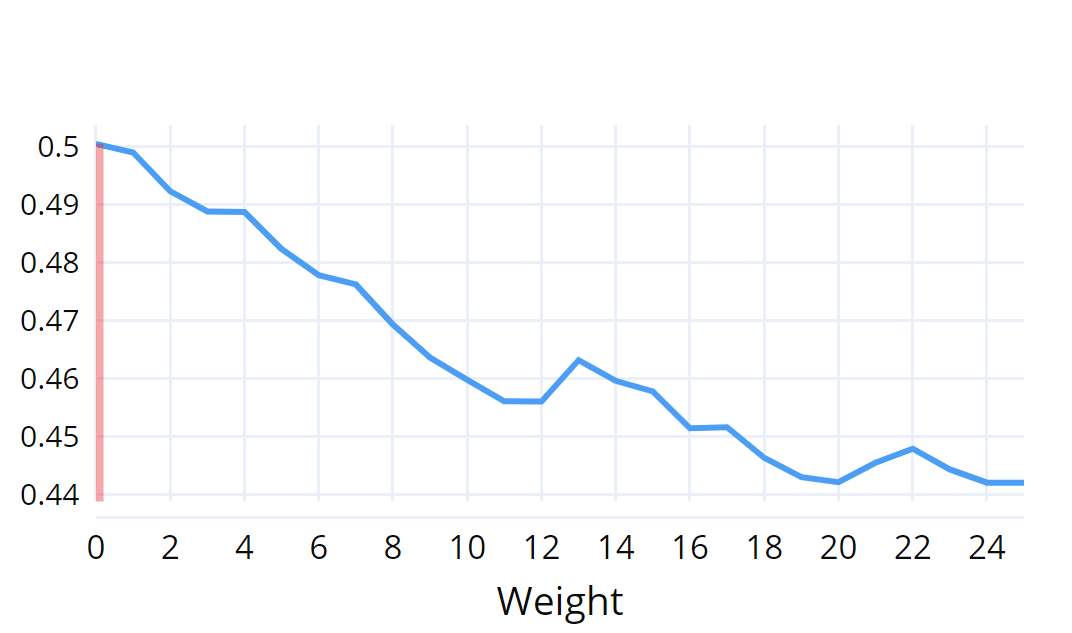}
        \caption{Similarity Scores of Doorways}
        \label{fig:sub4}
    \end{subfigure}
    \begin{subfigure}{.5\textwidth}
        \includegraphics[width=0.95\linewidth]{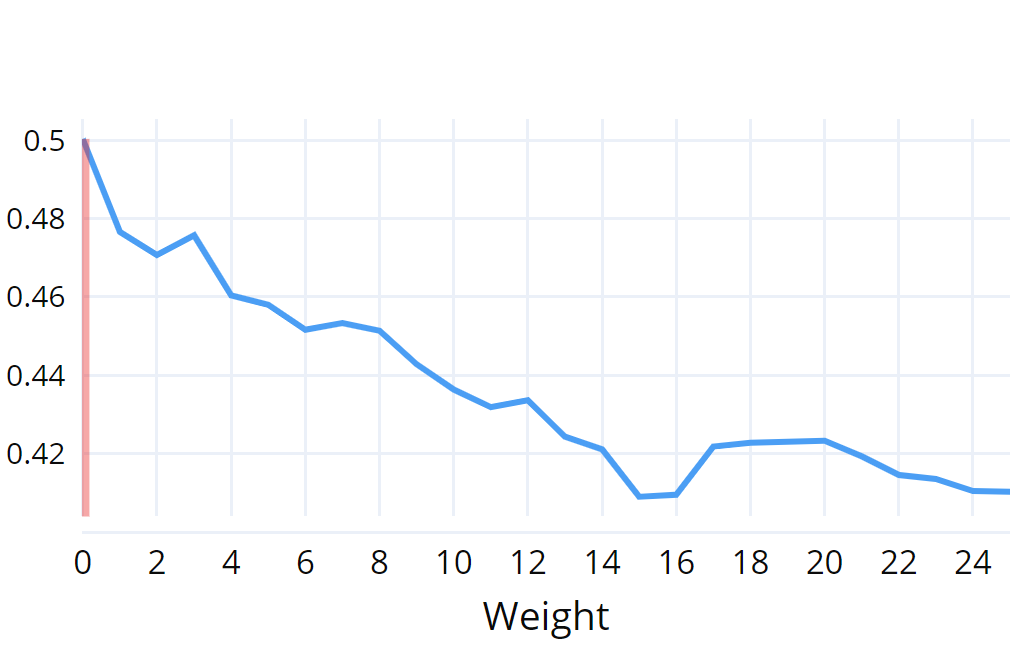}
        \caption{Similarity Scores of Linearity}
        \label{fig:sub5}
    \end{subfigure}%
    \begin{subfigure}{.5\textwidth}
        \includegraphics[width=0.95\linewidth]{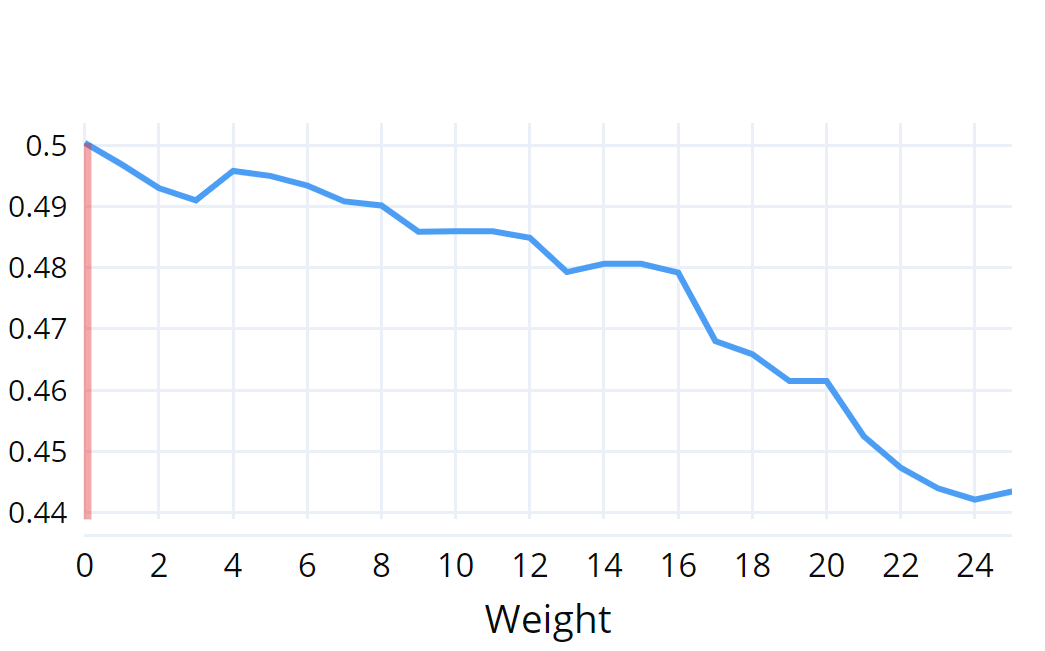}
        \caption{Similarity Scores of Staircases}
        \label{fig:sub6}
    \end{subfigure}
\caption{Similarity Scores of Noninfluential Criteria}
\label{fig:nic}
\end{figure}

We proceeded in the same way for each value of $w$ in the interval of $[0,25]$ (see Figure \ref{fig:nic}). For $25<w\leq 100$ we changed the step size to $10$. Should a higher similarity be observed here than was calculated for weights in range 0 to 25, the scope of weights that are sampled in intervals of 1 was extended. This served to ascertain that the best possible results did not lie beyond the boundary that was established.

\begin{figure}[tb]
\centering
    \begin{subfigure}{.5\textwidth}
        \includegraphics[width=0.95\linewidth]{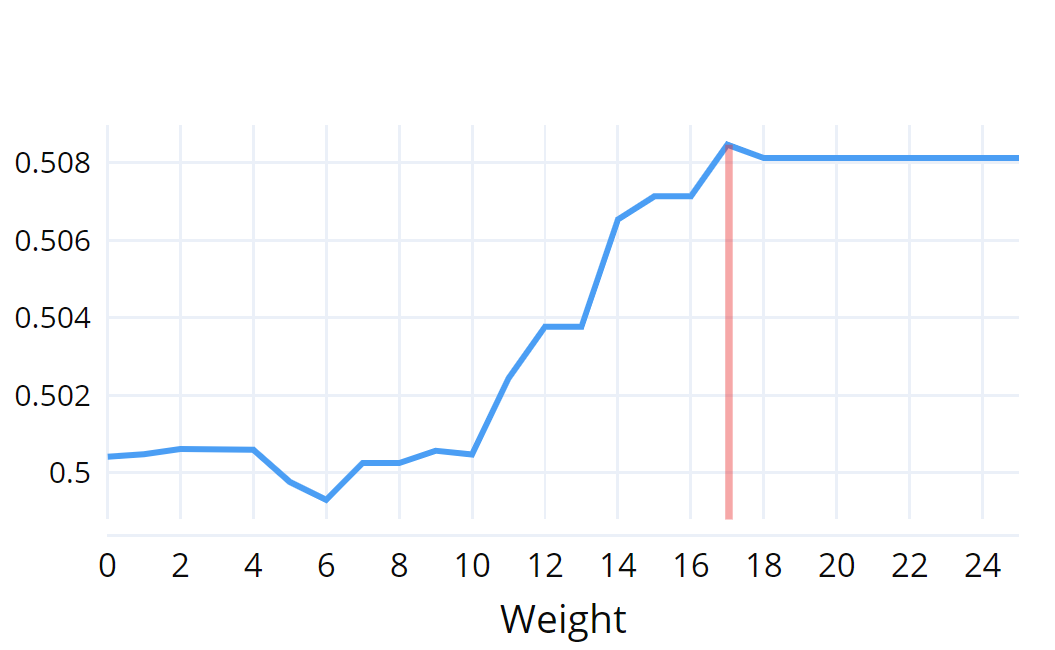}
        \caption{Similarity Scores of Elevators}
        \label{fig:sub7}
    \end{subfigure}%
    \begin{subfigure}{.5\textwidth}
        \includegraphics[width=0.95\linewidth]{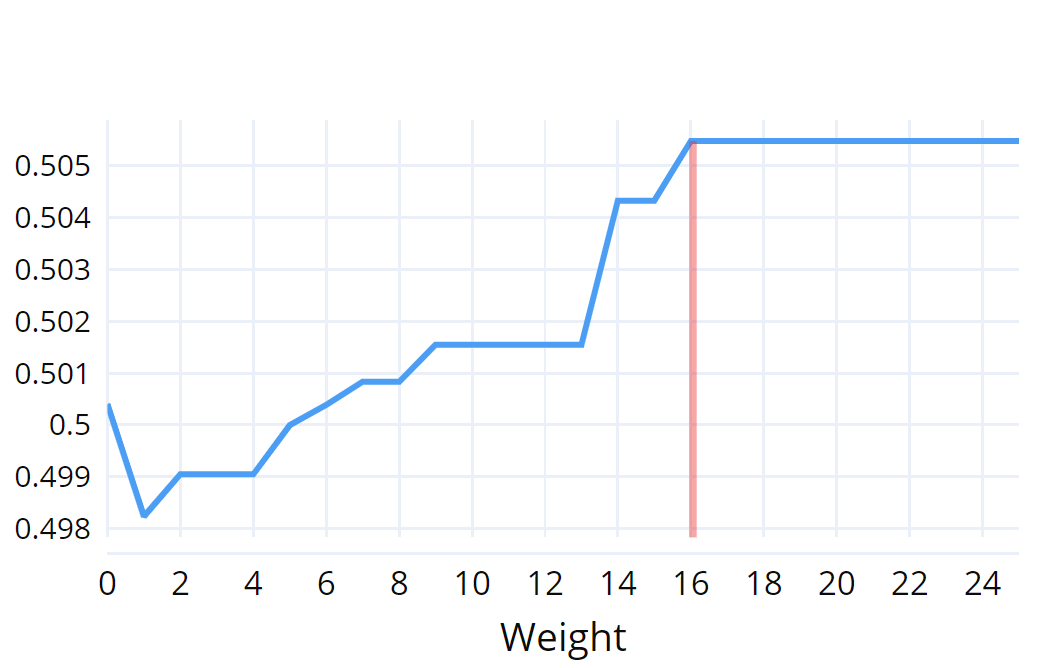}
        \caption{Similarity Scores of Entryways}
        \label{fig:sub8}
    \end{subfigure}
    \begin{subfigure}{.5\textwidth}
        \includegraphics[width=0.95\linewidth]{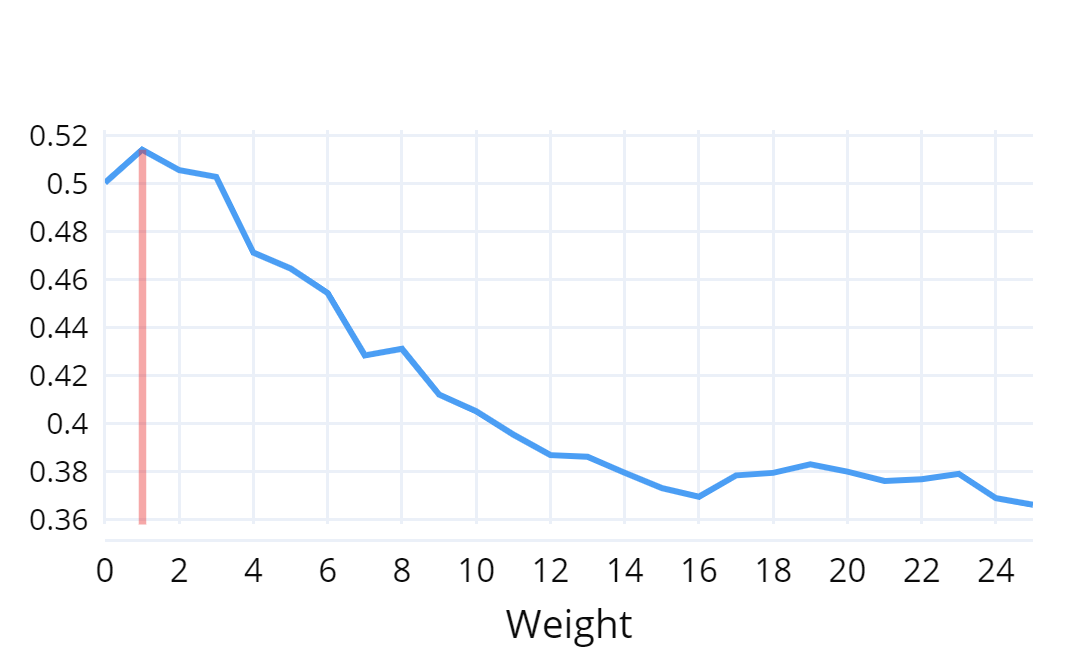}
        \caption{Similarity Scores of Streets}
        \label{fig:sub9}
    \end{subfigure}%
    \begin{subfigure}{.5\textwidth}
        \includegraphics[width=0.95\linewidth]{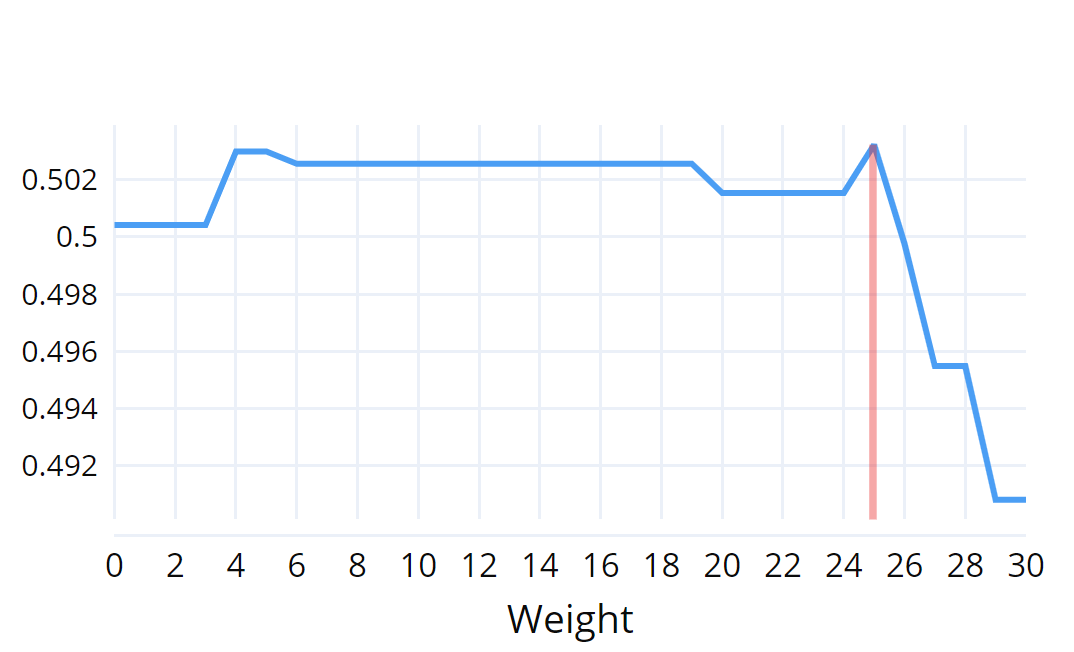}
        \caption{Similarity Scores of Revolving Doors}
        \label{fig:sub10}
    \end{subfigure}
    \begin{subfigure}{.5\textwidth}
        \includegraphics[width=0.95\linewidth]{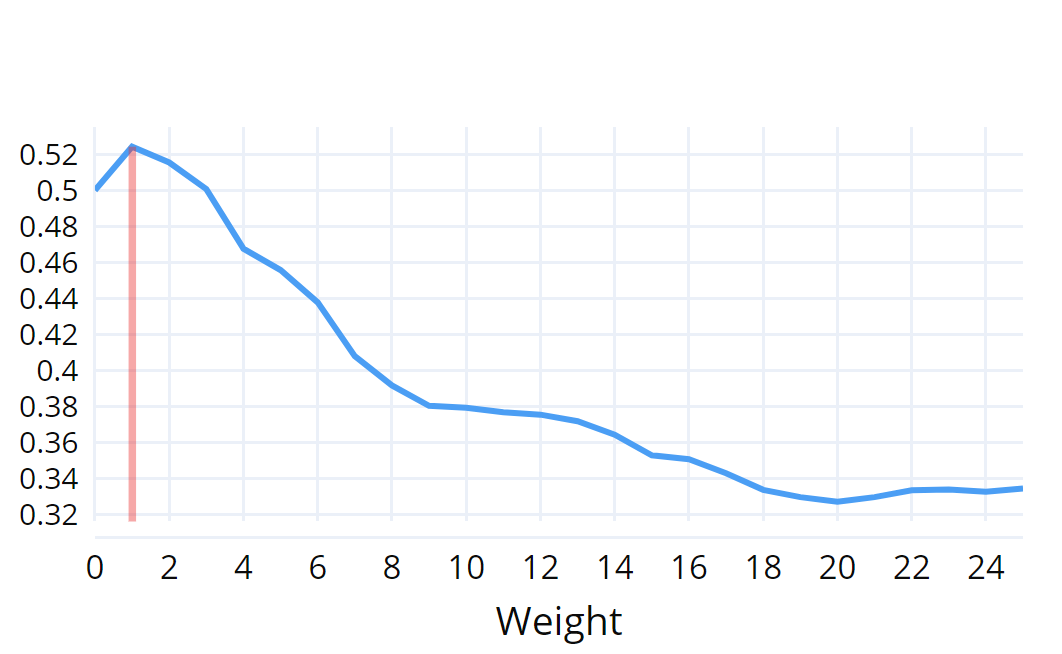}
        \caption{Similarity Scores of Turns}
        \label{fig:sub11}
    \end{subfigure}
\caption{Similarity Scores of Influential Criteria}
\label{fig:ic}
\end{figure}

The maximum detour from the shortest path in order to satisfy user preferences that would still be considered acceptable amounted to $w=100$ meters. However, as can be seen in Figures \ref{fig:nic} and \ref{fig:ic} the similarity between calculated paths and user preferred routes was maximal for $w<25$ for each of the factors considered.

\section{Results}
The work was conducted to find answers to the following questions, the answers to which are laid out in corresponding subchapters:
 \begin{enumerate}
     \item Does the consideration of the individual criteria, which were found to be influential to human wayfinding in previous work, lead to a higher similarity to paths that align with human preferences compared to the shortest path? And if so, to what extent?
     \item Assuming independence, how does a combination of all criteria with their respective optimal weights impact path similarity? 
    \item How do the criteria found to be influential compare to one another regarding their impact on human path selection?
 \end{enumerate}

\subsection{Individual Evaluation}

As described in the preceding chapter, for each factor, the average Similarity score was calculated in dependence on its weighting in order to determine which weight would lead to the best possible average approximation of the entire set of 221 user paths. Starting off the weighting with an increase of 1 in the range of 0 to 25, or if needed higher, provided the weight which leads to the optimum result. The unaltered shortest path displayed a similarity of 50\% to the user paths.

As shown in the following figure, not all factors led to an increase in similarity regardless of the added weight. This proved to be the case for the following factors: decision points, branching factors, doorways, staircases, minimum angle to the destination, and linearity at decision points. For the 5 Results 28 sake of a uniform presentation, all plots encompass a sampling of weights in the range of 0 to 25.

An improvement of the similarity means was therefore reached for all the other inspected criteria, which were turns, streets, entryways, revolving doors, and elevators. The plots depicted in the plots figure 2 visualize the relation between the weights and their improvement, a perpendicular red line marking the weight belonging to the corresponding maximum similarity score.

\subsection{Combination}
Assuming independence, all criteria that have been found to bring about a positive change in similarity score in the preceding section were then combined into one algorithm, each with their established best weight. \\
\indent
Due to the complementary nature of the factors turns and streets, which both penalize slightly altered variations of a deviation from a straight path and are thus evidently not independent from each other, taking into account both of them would lead to duplicate weights being added for some edges; only turns were therefore considered in this step since a bigger increase in similarity was observed. With an absolute similarity score of 52.30\%, this combination of criteria also resulted in a better approximation of user paths.

\subsection{Comparison}
Figure 3 lists the results for each factor and their combination in descending order of resemblance to the user paths measure by similarity score. Additionally, the percentage of affected paths is displayed.

\begin{figure}[tb]
\centering

\begin{center}
\begin{tabular}{ |c|c|c|c|c| } 
\hline
Factors & Weight    & Similarity Score  & Difference to SP & Impacted Paths\\
\hline
Turns & 1  & 52.47\%  & 2.43\%  & 19.91\% \\
Combination & varied  & 52.30\% & 2.26\% & 32.58\%  \\
Streets & 1  & 51.48\%  & 1.38\%  & 19.46\%  \\
Elevator & 17  & 50.85\%  & 0.81\%  & 8.60\%  \\
Entryway  & 16+  & 50.55\% & 0.51\%  & 11.76\% \\
Revolving Door & 25 & 50.32\%  & 0.28\%  & 2.26\% \\
 \hline
\end{tabular}

\caption{Sorted Table of Criteria with Scores and Percentage of Impacted Paths}

\end{center}
\end{figure}

The greatest increase in similarity measured 2.43\% and was achieved by turns, i.e. a reduction in the number of turns spanning 90 degrees or less, followed by the combination of criteria with 2.26\%. The avoidance of revolving doors produced the smallest improvement of 0.28\%.

\section{Discussion}
In consideration of the individual factors, one must retain that not in each case it was a higher number of occurrences that increased or decreased the assessed similarity to user paths. Following the elaboration of the criteria’s specific implementation in 4.3, it was their avoidance instead that was implied. \\
\indent
Many of the considered influential factors, previously found to be impactful, did not lead to the computation of paths that were more similar to those people prefer. This was the case for paths that integrated the avoidance of decision points, their reduced branching factors, the avoidance of entrances to rooms, as well as staircases, and a preference for choosing paths that either minimized the global angular distance to the destination’s location or aligned with the most linear path selection option.  Consequently, all other factors – the avoidance of elevators, entryways, revolving doors, turns, and streets – produced an increase in similarity.

A comparison shows that not only did the criteria’s impact on the similarity to user paths vary, but so did the number of affected paths: the differences between the “best” factor (turns) and “worst” one (revolving doors) are almost ten-fold regarding both similarity (2.43\% vs. 0.28\%) as well as impacted paths (19.91\% vs. 2.26\%). This certainly may be explained by the frequency with which each factor or entity occurs, for to produce significant changes it has to be present in the paths initially suggested by the shortest path algorithm.  \\
\indent
A preference for the most impactful factor can be reinforced by the finding that with an increasing number of turns included in a route, the likelihood of making a mistake while navigating, and thus the time needed for task completion, increases \cite{o1992effects}. The erroneous perception that routes containing more turns are longer than routes with fewer turns despite being equal in length (Dalton, 2003), might further explain our results. \\
\indent
In the cases of revolving doors, elevators, and entryways a boundless increase of their weights lead to an improvement (Figures X), signifying that the banishing of all occurring entities would be in the interest of users. Unsurprisingly, since a study conducted on part of the same data set already discovered a significant decline in the number of entryways in preferred routes \cite{muller2017path}, it was found that adding weights for all occurring entryways would improve the similarity. \\
\indent
The absence of entryways in the favored paths can be understood by considering the inconvenience this would imply, for taking a shortcut through lecture halls would generally only serve as an option during breaks, or passing through shops or offices would be equally unreasonable. A case against revolving doors can be made by considering that these types of doors are generally not barrier-free and users of wheelchairs would naturally avoid them. \\
\indent
The avoidance of choosing elevators as a means of transitioning to other floors implies that people favored the alternative option of taking stairs, which also concurs with previous findings \cite{kikiras2006ontology, muller2017path}. \\
\indent
The results found are much the same for turns and streets, which can be explained with the apparent similarities of the factors. Since turns only penalized right-angled deviations, and reducing the number of streets happened by penalizing all deviations from a straight path, reducing the number of streets can be viewed as an extension of turns since above 90 degrees (excluding 180 degrees) were also included.\\
\indent
Despite apparent similarities, a preference for linearity was not found and neither was a preference for minimizing the angular distance. These results could be attributed to decision points, which by themselves, and consequently their branching factor as well, were found ineffective. In addition, although such preferences have formerly been identified \cite{dalton2003secret, turner2009role}, the environments in which the studies took place differed greatly from navigating a university campus.\\
\indent
To find the optimal integration of multiple criteria into one algorithm, their specific weights would need to be identified anew. As was expected given the incorrectly assumed independence of the variables, the results fell short of reducing the number of turns alone. For taking dependencies between the factors into account, as a next step, we will train a predictor for $\mbox{wlength}(x,y)$ -- i.e.\ a neural network that allows non-linear function approximation.

Given all these limitations and observations from our analyses, we can still claim that our algorithm finds a good compromise between shortest paths and user preferred routes: The objective of minimizing path distance, which was evident in every route calculation performed, achieved conformity with the user paths of 50\% on average without further augmentation, and can therefore attest to align with human preferences. The overall results are encouraging for the incorporation of the criteria, which were found to lead to an increase of the similarity score, to create an algorithm able to suggest paths that humans like to follow.

\section{Conclusion}
This work aimed to acknowledge users as the focal point of navigational support systems and therefore suggest paths that match with their preferences. To quantify the extent to which this is accomplished, a set of paths, which people take in their daily lives, was considered representative of human preferences, and similarity to these paths was treated as a measure of accomplishment. The assumption that people enjoy being suggested the same routes that they take in well-known environments might be flawed, since aspects regarding the navigation app, such as the ease with which instructions can be displayed and comprehended, are disregarded. Some of the factors were not found to be influential in an indoor environment specifically, but rather in settings whose similarity to the one considered here seems dubious. Narrowing down the findings of previous studies to research applicable to an indoor setting might help identify influential factors more successfully. On the upside, putting these findings into a different context sheds light on their transferability and may offer insights into why they have been found to impact wayfinding decisions. Future research might proceed by taking criteria into consideration that have not been discussed and considered here, including not only environmental factors but more subjective criteria which might be more difficult to assess. Lastly, shedding more light on the interrelations of influential factors might be necessary for the conception of an ideal algorithm that can consider more than one element at a time.

As far as the comparison between indoor and outdoor environments is concerned, our study reproduces the results reported in \cite{doi:10.1080/13658816.2010.510799} for indoor environments: user prefer routes with a small number of turns. However, as indoor environments are more complex that streets outside, also other criteria are influential. It is an interesting challenge for future work whether a transformation of a graph in which nodes actually represent actual distances physical properties of the environment to another model in which nodes represent inevitable choices (such as taking a turn) is feasible if multiple criteria have to be considered simultaneously.


\bibliography{paper_11}

\appendix

\end{document}